\documentclass[runningheads]{llncs}

\usepackage{eccv}

\usepackage{eccvabbrv}
\usepackage[symbol]{footmisc}
\usepackage{graphicx}
\usepackage{booktabs}
\usepackage{adjustbox}
\usepackage{subcaption}
\usepackage{wrapfig}
\usepackage{overpic}
\usepackage[accsupp]{axessibility}  

\usepackage{hyperref}

\usepackage{orcidlink}
\usepackage{multirow}
\usepackage{multicol}
\usepackage{algorithm}
\usepackage{algorithmic}
\newcommand{\methodname}{RAM}
\newcommand{\bonus}[1]{{\tiny\textbf{\color{red}#1}}}
\newcommand{\decline}[1]{{\tiny\textbf{\color{blue}#1}}}
\definecolor{codegray}{rgb}{0.55,0.57,0.55}
\definecolor{mygray}{gray}{.9}
\usepackage{colortbl}
\begin{document}
\title{Restore Anything with Masks:  Leveraging Mask Image Modeling for Blind All-in-One Image Restoration} 

\titlerunning{Restore Anything with Masks}

\author{
Chu-Jie Qin\inst{1,2}\thanks{A part of this work is done during Chu-Jie Qin's internship at Samsung.} \and
Rui-Qi Wu\inst{1,2} \and 
Zikun Liu\inst{3} \and
Xin Lin\inst{5} \and \\
Chun-Le Guo\inst{1,2} \and
Hyun Hee Park\inst{4} \and
Chongyi Li\inst{1,2}\thanks{Chongyi Li is the corresponding author.} 
}

\authorrunning{Qin et al.}

\institute{VCIP, CS, Nankai University  \and
NKIARI, Shenzhen Futian \\
\email{\{chujie.qin,wuruiqi\}@mail.nankai.edu.cn}\\
\email{\{guochunle, lichongyi\}@nankai.edu.cn}
\and
Samsung Research, China, Beijing (SRC-B)\and
The Department of Camera Innovation Group, Samsung Electronics \\
\email{\{zikun.liu,inextg.park\}@samsung.com} \and
Sichuan University  \\
\email{linxin@stu.scu.edu.cn}\\
}
\maketitle
\begin{abstract}
 All-in-one image restoration aims to handle multiple degradation types using one model. 
 This paper proposes a simple pipeline for all-in-one blind image restoration to \textbf{R}estore \textbf{A}nything with \textbf{M}asks (\textbf{\methodname}).
 We focus on the image content by utilizing Mask Image Modeling to extract intrinsic image information rather than distinguishing degradation types like other methods.
 Our pipeline consists of two stages: masked image pre-training and fine-tuning with mask attribute conductance. We design a straightforward masking pre-training approach specifically tailored for all-in-one image restoration. This approach enhances networks to prioritize the extraction of image content priors from various degradations, resulting in a more balanced performance across different restoration tasks and achieving stronger overall results. To bridge the gap of input integrity while preserving learned image priors as much as possible, we selectively fine-tuned a small portion of the layers. Specifically, the importance of each layer is ranked by the proposed Mask Attribute Conductance  (\textbf{MAC}), and the layers with higher contributions are selected for finetuning. Extensive experiments demonstrate that our method achieves state-of-the-art performance. Our code and model will be released at \href{https://github.com/Dragonisss/RAM}{https://github.com/Dragonisss/RAM}.
\end{abstract}    
\section{Introduction}
\label{sec:intro}
Image restoration involves the restoration of low-quality images affected by various degradation, typically arising from adverse environmental conditions (\eg, rain, haze, low-light), hardware-related issues (\eg, noise and blur), and post-processing artifacts (\eg, JPEG compression).
Image restoration serves not only to enhance the visual appeal of images but also contributes to practical application scenarios such as autonomous driving and surveillance.

\begin{wrapfigure}{r}{0pt}
    \centering
    \includegraphics[width=0.46\linewidth]{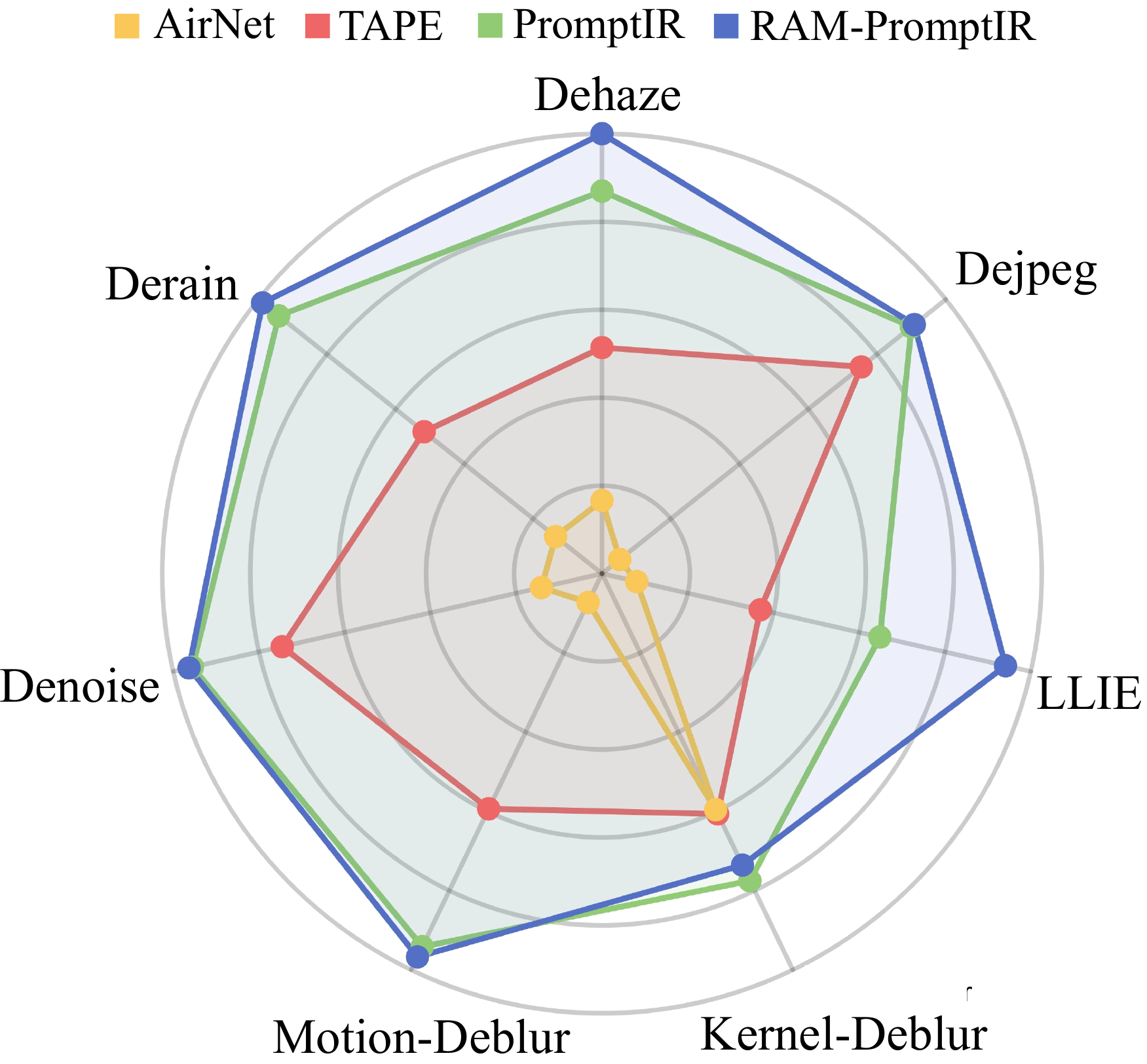}
    \caption{Our \text{RAM} achieves more balanced and more powerful performance than the state-of-the-art methods (AirNet~\cite{airnet}, TAPE~\cite{liu2022tape}, PromptIR~\cite{promptir}) for all-in-one blind image restoration.}
    \label{fig:teaser}
\end{wrapfigure}
Modern techniques in this field mainly focus on learning fixed patterns formed during the degradation process, \ie, degradation priors.
Some works~\cite{rain_1, rain_2, lowlight2} utilize task-specific priors to solve a certain degradation problem,
while another research line~\cite{liang2021swinir,Zamir2021Restormer,mehri2021mprnet,nafnet,wang2022uformer} tries to design a general network architecture that can effectively learn each degradation pattern.
Nevertheless, the above methods only enable the network to learn a single degradation, resulting in an imbalanced situation when dealing with multiple types of degradation.

To tackle the problem stated above, all-in-one methods have emerged, aiming to handle multiple degradations using one model.
Most of these approaches tend to utilize explicit priors (\eg, AirNet~\cite{airnet}) or introduce an extra module (\eg, PromptIR~\cite{promptir}) to discern image degradation patterns, thereby assisting the model in performing the restoration.
However, these methods place their emphasis on distinguishing degradation types in images rather than the image content, leading to lower scalability and fuzzy decision boundaries when more degradation types are involved. 
We argue that the essence of image restoration is to extract intrinsic image information from corrupted images rather than eliminate degradation patterns, \ie, learning image prior rather than degradation prior.
It is worth noting that TAPE~\cite{liu2022tape} similarly suggests that understanding normal image nature aids restoration by introducing a natural image prior.
Nevertheless, TAPE utilizes the model output as the optimization target, which causes the model to amplify its own errors and learn the image prior with bias.

In this paper, we focus on tackling \textbf{how to extract intrinsic image information from diverse corrupted images.} 
Some attempts~\cite{mim_lowlevel,mim_denoising} by Mask Image Modeling (MIM) in low-level vision have caught our attention. As a pre-training strategy, MIM has been widely validated for its effectiveness in high-level tasks, thanks to its generic representation of images. 
Simultaneously, the model also learns the distribution of natural images, which encompasses the intrinsic information we aim to extract from the images.
Built on MIM, we propose a simple pipeline for all-in-one blind image restoration that \textbf{R}estores \textbf{A}nything with \textbf{M}asks (\textbf{\methodname}), which includes two stages: the mask pre-training stage and the fine-tuning stage with Mask Attribute Conductance (MAC). In the pre-training stage, we randomly mask corrupted images at the pixel-wise level and force the network to predict the clear one corresponding to the masked pixels, extracting inherent image information from corrupted images. In the fine-tuning stage, we focus on overcoming the input integrity gap caused by changing masked input during pre-training into the whole image during inference while preserving learned prior as much as possible. 

Specifically, we first evaluated the importance of each network layer in addressing this gap by the proposed MAC. Following that, we chose the top $k\%$ most critical layers for fine-tuning while keeping the rest of the network layers frozen. We demonstrate that after a brief fine-tuning period (even if only $10\%$ layers are tuned), the model can achieve a highly satisfactory performance level,  surpassing models trained using traditional pair-wise training. 
Additionally, our pipeline can be plug-and-play used in any network without introducing additional computational overhead.

The contributions of this work are as follows:
\begin{itemize}
\item We discuss the challenge of adopting MIM in low-level vision and propose a MIM-based pre-training strategy tailored to all-in-one blind image restoration, which allows the restoration networks to effectively learn inherent image information while guaranteeing reconstruction results.
\item 
We proposed Mask Attribute Conductance to evaluate the importance of each layer in addressing the input integrity gap so that a very small portion (\eg $10\%$) of critical layers are tuned to bridge this gap while preserving the image prior learned by MIM.
\item  Our proposed RAM provides a fresh perspective to achieve more balanced and powerful all-in-one blind image restoration, which focuses on extracting inherent image information from corrupted images. 
Our pipeline can be applied to any image restoration network without introducing additional computational overhead.
\end{itemize}

\section{Related Work}
\label{sec:related_work}
\begin{figure*}[t]
  \centering
   \includegraphics[width=0.95\linewidth]{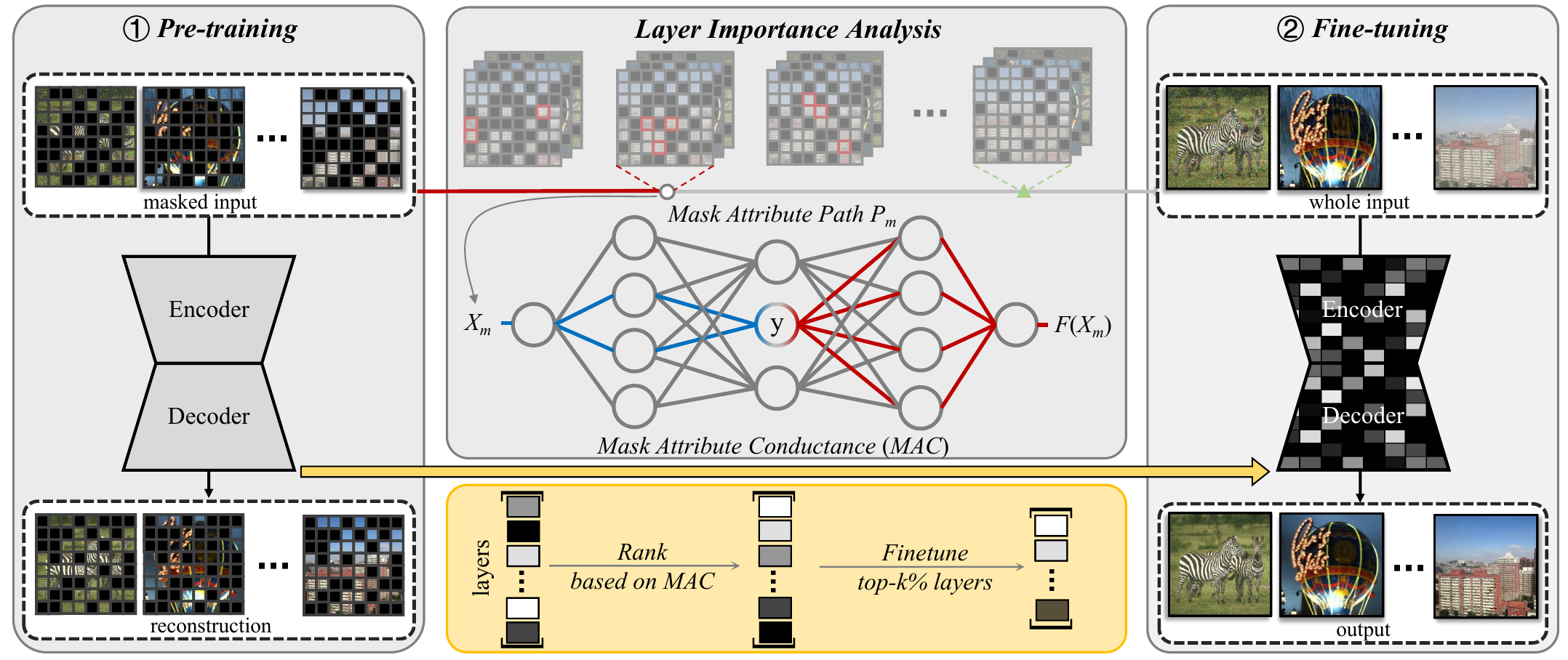}
   \caption{The illumination of our overall pipeline. 1) Pre-training the model with mask image pre-training method tailored to low-level vision. We randomly mask degraded images at the pixel level with a $50\%$ masking ratio and reconstruct the clean images. 2) The Fine-tuning stage is followed to overcome the input integrity gap caused by changing masked input during pre-training into the whole image during inference.
    We analyze the importance of each network layer for resolving the input integrity gap according to the proposed MAC and rank them in descending order. The top $k\%$ of network layers are selected for fine-tuning on the complete image. 
    }
   \label{fig:overall pipeline}
\end{figure*}
\subsection{Image Restoration for Multi Degradations}
While neural networks have demonstrated impressive performance in single degradation image restoration~\cite{deblur_1,deblur_2,guo2020zerodce,guo2022dehamer,wu2023ridcp,lowlight2,lowlight3,rain_1,rain_2,jin2023dnf}, recent works have shifted their focus towards addressing the more challenging domain of multi-degradation image restoration. 
A group of methods~\cite{nafnet,wang2022uformer,Zamir2021Restormer,liang2021swinir,mehri2021mprnet} aims at designing a general architecture that can effectively learn each degradation pattern.
SwinIR~\cite{liang2021swinir} employs a window attention mechanism to convert global attention into a localized approach, effectively reducing computational overhead.
In addition, the U-shaped transformer-based methods~\cite{wang2022uformer,Zamir2021Restormer} are employed to extract multi-scale features and reduce computational overhead.
However, these methods have to train individually on each restoration task.
Several methods \cite{li2020all,ipt} leverage multiple input and output heads to empower the network to restore various types of degraded images. Nonetheless, this kind of approach may lead to the diminished scalability of the model.
Recently, several subsequent methods \cite{airnet,two,degradation,promptir,expert,men,park} have been proposed to employ a unified network to address multiple restoration issues.
Most of these methods put emphasis on learning how to distinguish different types of degradations and restore corrupted images.
Typically, AirNet \cite{airnet} first proposed an all-in-one image restoration task. The method initially pretrains a degradation classifier based on contrastive learning and subsequently utilizes it to assist in all-in-one image restoration.
PromptIR \cite{promptir} has introduced a learnable prompt-based module. Instead of constraining the degradation category, it enables the model to autonomously learn features that are advantageous to its performance by using an adaptive prompt.
Our \methodname{} takes a fresh perspective that focuses on extracting common content information from corrupted images, without any extra design to distinguish degradations,
which helps us achieve balance and powerful performance when more degradation types are taken into consideration.
\subsection{Mask Image Modeling}
Inspired by Mask Language Modeling~\cite{bert,gpt2018}, Mask Image Modeling (MIM)~\cite{mae,simmim} is introduced as a pretraining approach to learn general representations in high-level vision. MAE \cite{mae} effectively utilizes MIM for predicting hidden tokens, demonstrating strong performance and generalization across various downstream tasks.
SimMIM \cite{simmim} proposed a general masked image modeling method based on Swin-ViT \cite{swin}.
Painter \cite{painter} unifies multiple tasks under image-to-image translation and leverages MIM pretraining.
In recent years, there have been efforts to incorporate MIM into the realm of low-level vision to enhance model generalization.
Among them, \cite{mim_denoising} and \cite{mim_lowlevel} are the most closely aligned with our focus.
\cite{mim_denoising} employs the MIM model to enhance the model's generalization for denoising tasks but has not explored its potential in multi-task scenarios.
\cite{mim_lowlevel} utilizes MIM for pre-training the model encoder to introduce generative prior and subsequently employs the decoder for restoration. However, it does not fully harness the potential of MIM. 
Our proposed RAM utilizes MIM to unify the optimization objective for various image restoration tasks into reconstructing intrinsic image information. This allows the network to learn restoration functions more balanced and effectively. Moreover, to preserve the image priors learned by MIM, we designed a fine-tuning strategy based on MAC analysis (in ~\cref{sec:finetuning_stage}). This enables us to achieve comparable performance by fine-tuning only a small portion (\eg $10\%$) of layers, fully tapping into the potential of MIM.


\subsection{Gradient-based Attribution}
 Gradient-based attribution methods~\cite{ig,approx_ig,layerconductance,shrikumar2018computationally,lam,faig} are often used to clarify how hidden units (or inputs) impact the output of networks.
 One commonly used approach is Integrated Gradients (IG) \cite{ig,approx_ig}, which accumulates gradients along a linear path from the baseline input to the target input in the pixel/feature space. After that, IntInf \cite{intinf} and layer conductance \cite{layerconductance} alter IG to attribute neuron importance along the same path. 
 In our work, we expect to find the key layers that can effectively overcome the distribution shift between training data and inference data. We propose Mask Attribute Conductance (MAC) based on the layer conductance and accumulated MAC of each layer along the Mask Attribute Path (MAP). MAC can represent the layer's importance along the MAP. In this way, we can fine-tune the top $k\%$ critical layers of the pre-trained network, preserving to a great extent the image priors learned during pretraining.

\section{Methodology}
In this section, we start with discussing the challenges of using MIM in low-level vision tasks (\cref{sec:mim_discussion}). Following that, we present our pipeline for all-in-one blind image restoration, which contains two parts: pre-training with MIM (\cref{sec:pretraining_stage}) and fine-tuning with Mask Attribute Conductance (MAC) Analysis (\cref{sec:finetuning_stage}).
\subsection{Rethinking MIM in Low-Level Vision}
\label{sec:mim_discussion}
MIM is a process that randomly masks certain parts of an image and extracts features from the remaining visible parts to reconstruct the entire image.
It allows models to acquire a generic representation of images and thus achieve good pre-training, which is verified in many high-level tasks~\cite{mae,simmim}.
Moreover, the models also learn the distribution of natural images during the image reconstruction, \ie MIM pre-training.
This incidental acquisition of prior knowledge is instrumental in tasks like image restoration.
Despite these advantages, applying MIM in pretraining a model for low-level vision tasks is still under-explored, primarily due to the challenges that must be addressed in the process.

\begin{figure}[t]
    \centering
    \includegraphics[width=\linewidth]
    {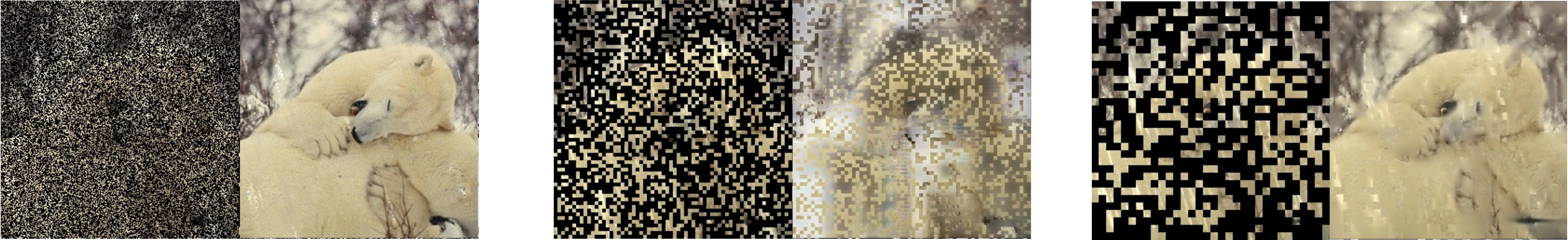}
    \put(-320,-10){patch size=1}
    \put(-200,-10){patch size=4}
    \put(-80,-10){patch size=8}
    \caption{Mask Image Modeling reconstruction with different patch sizes. We pre-trained with different patch sizes and visualized the mask inputs (left), and the corresponding MIM reconstructions (right).}
    \label{fig:patch-size}
\end{figure}

Firstly, the main purpose of vanilla MIM is not high-quality reconstruction but good feature extraction for high-level tasks.
Therefore, it masks a wider range of images to gather semantic information but not pixel-level content, reflected in token-level masking and a high mask ratio. CSFormer \cite{mim_lowlevel} directly adopts this strategy on low-level vision pre-training. However, some studies verify that semantic information is not as important for image restoration as it is in pattern recognition tasks~\cite{liu2021discovering, magid2022texture}. Moreover, high-degree masking leads to producing detail-deficient results, as shown in \cref{fig:patch-size}, which is harmful to low-level tasks.

Secondly, the training objective of MIM is to reconstruct the masked input images, so it can only produce results with the same domain as the input image. However, we hope the model gains the ability to bridge low-quality domain to high-quality domain, \ie recover clean content from degraded input.
Therefore, it is necessary to introduce paired data when pre-training image restoration models by MIM (see the experiment in~\cref{sec:ablation} for details). 
Chen \etal \cite{mim_denoising} demonstrate that pair-wise MIM training enhances the generalization performance over different types of noisy images. In this paper, we take a step forward to explore the effectiveness of MIM on multiple degradations with larger variance.

\subsection{Pretraining with MIM}
\label{sec:pretraining_stage}

Based on the above analysis, we design a MIM pre-training paradigm tailored for low-level vision.

\noindent \textbf{Masking.} During the pre-training stage, we randomly mask the pixels of degraded images (mask images in a $1 \times 1$ patch size) with a $50\%$ mask ratio.
We found that fine-grained masked patches and balanced mask ratio are beneficial to image restoration, which can be demonstrated in Sec.~\ref{sec:ablation}.

Besides, since our MIM pre-training has a similar target to subsequent low-level tasks, we do not need to change the decoder like MAE~\cite{mae} does but just fine-tune it.

\noindent \textbf{Reconstruction target.} Following the Bert~\cite{bert} and MAE~\cite{mae}, we choose L1 loss to supervise the masked part. The training objective can be written as:
\begin{equation}
    \mathop{\arg\min}\limits_\theta\mathbb{E}[||\tilde{\mathcal{M}}(I- f(\mathcal{M}(I_d), \theta))||],
\end{equation}
where $\{I, I_d\}$ represents a pair of clean image and degraded image, $f(\cdot, \theta)$ denotes a network with parameters $\theta$, $\mathcal{M}(\cdot)$ is a random binary masking operation and $\tilde{\mathcal{M}}(\cdot)=1-\mathcal{M}(\cdot)$.
\begin{figure}[t]
  \centering
   \includegraphics[width=0.8\linewidth]{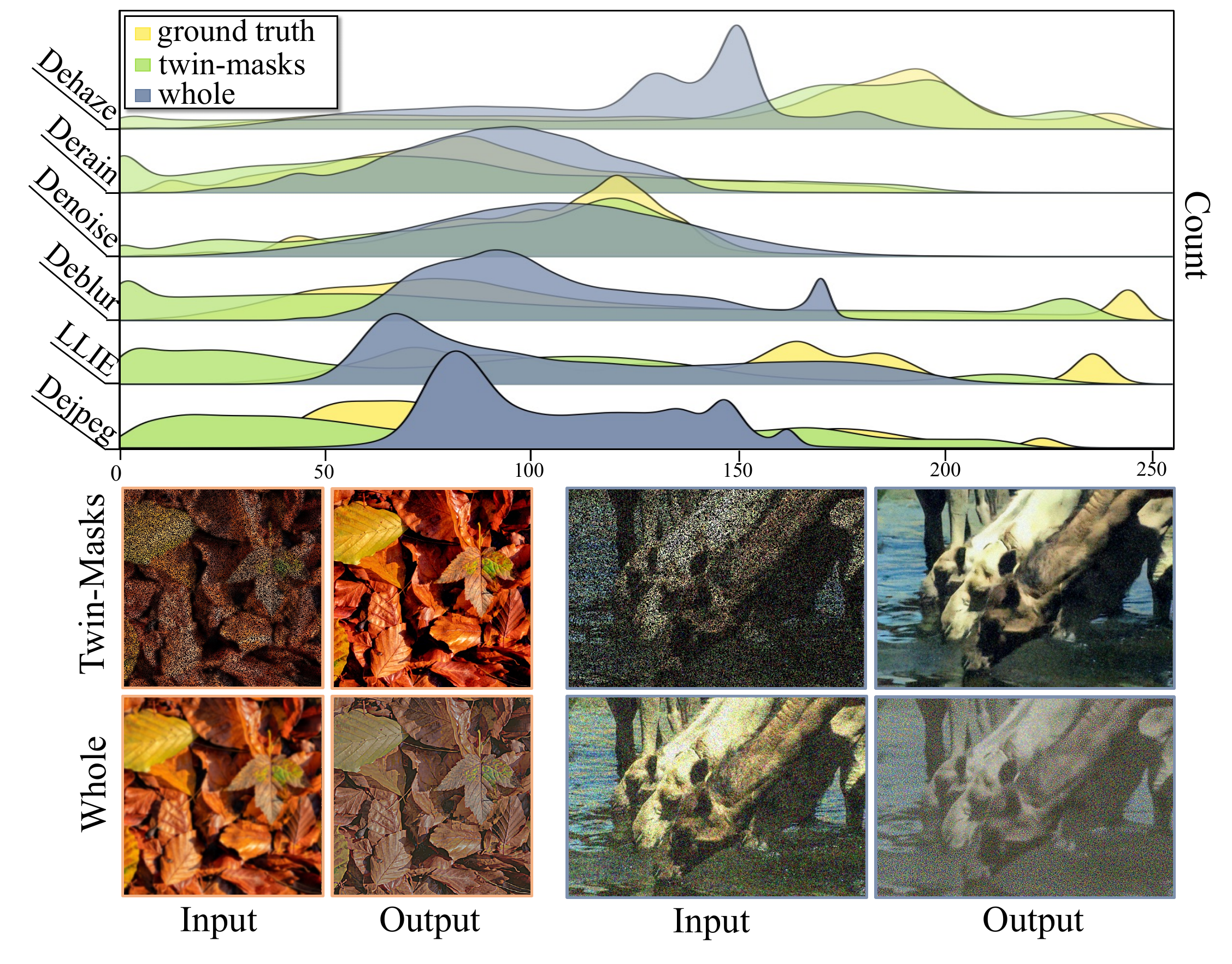}
   \caption{The effect of MIM reconstruction with different input integrity on kernel deblurring (orange border) and denoising (blue border). We also visualize the color distributions of reconstructions in various tasks above. It shows that the distribution of the reconstruction results obtained using the twin-masks method as input is closer to the real images (ground truth) compared to the results obtained using the whole input.}
   \label{fig:gap observation}
\end{figure}

\subsection{Finetuning with Mask Attribute Conductance Analysis}
\label{sec:finetuning_stage}

\noindent\textbf{Observation.} During pre-training, the network learns rich content priors. However, the incompleteness of the masked input prevents the direct use of the pre-trained model for inference, as it would result in a distribution shift in the outputs. As shown in \cref{fig:gap observation}, We start by feeding the entire image into a pre-trained model, leading to a color-distorted result. Next, we use a pair of complementary masks, referred to as twin-masks, to individually mask the image. Subsequently, we input both of these complementarily masked images into the network. By combining the pixel values predicted by each image, we generate a higher-quality image. 
This observation indicates that the hindrance to using mask pre-trained model directly for inference lies in input incompleteness rather than the model's inability to learn the restoration function.

Building upon this insight, we explore the possibility of minimizing the influence of disparities in data input formats via model fine-tuning. To maintain the learned priors, it is essential to retain pre-trained parameters as extensively as possible while employing the fewest but most effective layers for fine-tuning. To tackle this, we introduce the concept of mask attribution conductance, which quantifies the importance of each layer concerning the fine-tuning objective. We then identify the top-k\% most critical layers for fine-tuning. 

\noindent\textbf{Preliminary.} Before giving the definition of Mask Attribute Conductance (MAC), we briefly recall the definition of integrate gradient\cite{ig} (IG) and neuron conductance\cite{layerconductance} (Cond). Considering a linear path $\gamma(\alpha)=x'+\alpha(x-x')$ from base input $x'$ to target input $x$, we can attribute output change $F(x)-F(x')$ to $i$-th dimension of input/feature $x_{i}$ (\eg a pixel) by calculating its integrate gradient, which formally as below:
\begin{equation}
  \mathrm{IG}_{i}(x):= (x_{i}-x'_{i})\cdot\int_{0}^{1} \frac{\partial F(x'+\alpha(x-x'))}{\partial x_{i}}\, d\alpha.
  \label{eq:ig}
\end{equation}
We can also attribute output change to a specific neuron $y$ by improving IG, which involves calculating the conductance. The conductance\cite{layerconductance} of the hidden neuron $y$ along the $\gamma (\alpha)$ is:
\begin{equation}
\begin{split}
  \mathrm{Cond}^{y}(x
  )&:= \sum_{i}(x_{i}-x'_{i})\cdot\int_{0}^{1} \frac{\partial F(x'+\alpha(x-x'))}{\partial y}\cdot \frac{\partial y}{\partial x_{i}}\, d\alpha \\ 
  & = \sum_{i}\int_{0}^{1} \frac{\partial F(\gamma(\alpha))}{\partial y}\cdot \frac{\partial y}{\partial \alpha}\, d\alpha,
  \label{eq:cond}
\end{split}
\end{equation}
Note that $(x_{i}-x'_{i})=\frac{\partial (x'+\alpha(x_{i}-x'_{i}))}{\partial \alpha}$.
Certainly, we can broaden \cref{eq:cond} to compute conductance when integrating along any given path $\alpha:[s,t]\rightarrow P$:
\begin{equation}
  \mathrm{GeneralCond}^{y}(x):= \sum_{i}\int_{P} \frac{\partial F(X_{i}(\alpha))}{\partial y}\cdot \frac{\partial y}{\partial \alpha}\, d\alpha,
  \label{eq:generalcond}
\end{equation}

\noindent{}where $X:R\rightarrow R^{n}$ is the function of the path from $x'$ to $x$, which satisfies $X(s)=x'$, $X(t)=x$. $[s,t]$ represent the domain of the path function X." 

\begin{figure}[t]
  \centering
   \includegraphics[width=0.8\linewidth]{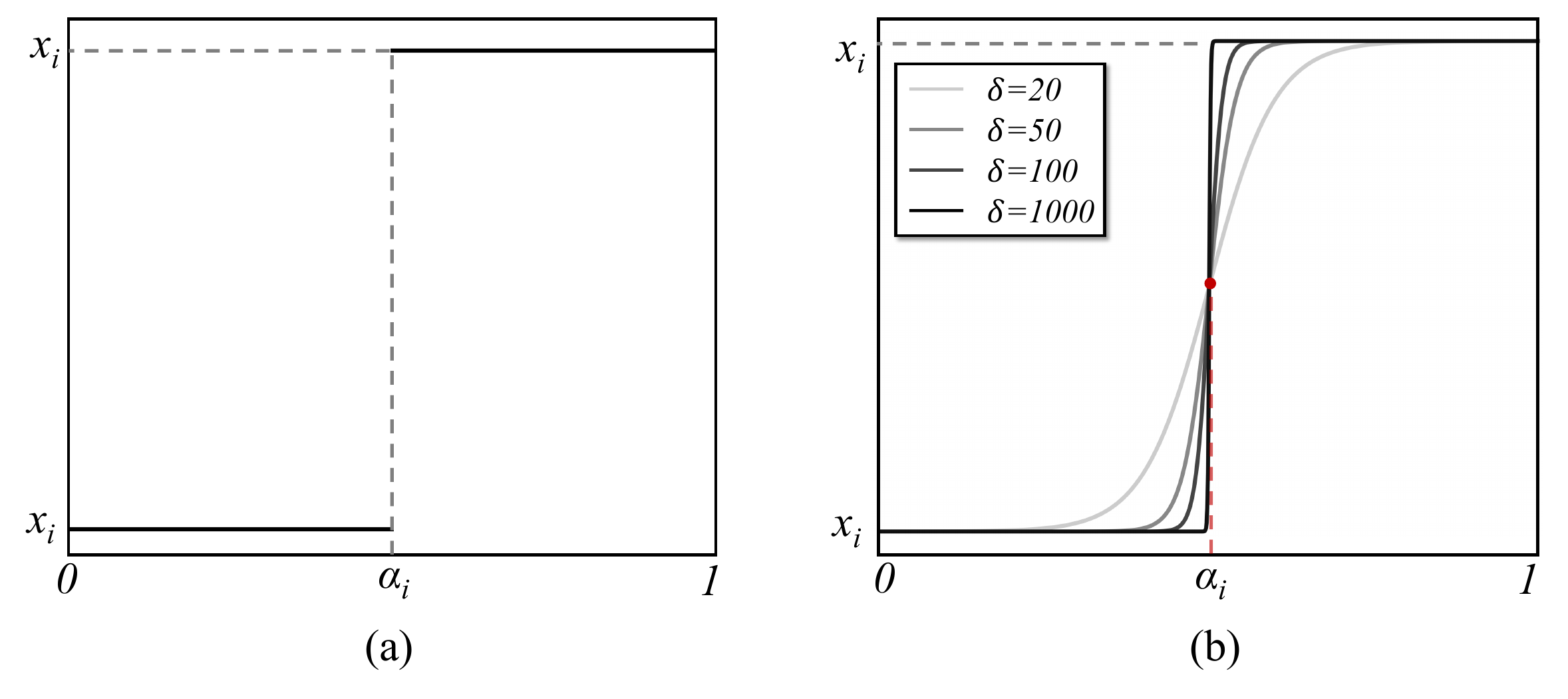}
   \caption{Illumination of (a) $X_i^{m}$ in \cref{eq:hardmaskpath} and (b) $\tilde{X}_{i}^{m}$ in \cref{eq:neuron softmaskpath}. }
   \label{fig:function}
\end{figure}

\noindent\textbf{Finetuning with MAC.} To find effective layers to finetune, we propose \textbf{M}ask \textbf{A}ttribute \textbf{C}onductance (\textbf{MAC}) to evaluate how effective each layer is in overcoming the gap of input integrity. Considering such a nonlinear path $\alpha: [0,1] \rightarrow P_{m}$ from zero input $x'$ to whole input $x$, which path function $X^m$ satisfies:
\begin{equation}
  X_{i}^{m}(\alpha;\alpha_i) = \begin{cases}
x'_{i}, & \alpha < \alpha_i \\
x_i, & else \\
\end{cases},
  \label{eq:hardmaskpath}
\end{equation}

where $i$ refers to the index of pixels, $\alpha_i\in (0,1]$ is a set of parameters that indicate when each pixel gets masked. We define this path as a Mask Attribute Path (MAP). Apparently, $X^m(0)=x'$ and $X^m(1)=x$. 

However, $X^m$ is not differentiable, making it an invalid attribute path function. To solve this problem, we use a group of sigmoid-like functions  $\tilde{X}^{m}$ to approximate $X^m$:
\begin{equation}
  \tilde{X}_{i}^{m}(\alpha;\alpha_i) = \frac{(x'_{i}-x_{i})}{1+e^{-\delta(x'_{i}-\alpha_{i})}}.
  \label{eq:neuron softmaskpath}
\end{equation}

We can see that $\tilde{X}^{m}$ is very close to $X^{m}$ when $\delta$ is sufficiently large (as depicted in~\cref{fig:function}). And for each $\tilde{X}_{i}^{m}$, it will change sharply from $x'_{i}$ to $x_i$ when $\alpha$ is in the neighborhood of $\alpha_{i}$.

Here, we can give a definition of \textbf{MAC} as below:
\begin{equation}
\begin{split}
  \mathrm{MAC}^{y}(x)&:= \sum_{i}\int_{P_m} \frac{\partial F( {X}_{i}(\alpha))}{\partial y}\cdot \frac{\partial y}{\partial \alpha}\, d\alpha \\
  & \approx  \sum_{i}\int_{0}^{1} \frac{\partial F( \tilde{X}_{i}^{m}(\alpha;\alpha_i))}{\partial y}\cdot \frac{\partial y}{\partial \alpha}\, d\alpha.
  \label{eq:mac}
\end{split}
\end{equation}

In fact, a partial path is also available to attribute from a masked input $x_m$ with any mask ratio $r$ to whole input $x$:
\begin{equation}
  \mathrm{MAC}_{r}^{y}(x)\approx  \sum_{i}\int_{1-r}^{1} \frac{\partial F( \tilde{X}_{i}^{m}(\alpha;\alpha_i))}{\partial y}\cdot \frac{\partial y}{\partial \alpha}\, d\alpha.
  \label{eq:mac_extend}
\end{equation}

In practice, we use N-steps discretization to approximate the integral form of \cref{eq:mac_extend}, which follows 
\cite{shrikumar2018computationally}:
\begin{equation}
\begin{split}
  \mathrm{MAC}_{r}^{y}(x) &\approx \sum_{i}\sum_{j=1}^{N} \frac{\partial F( \tilde{X}_{i}^{m}(\frac{jr}{N};\alpha_i))}{\partial y} \\
  &\cdot (F_{y}(\tilde{X}_{i}^{m}(\frac{(j+1)r}{N}))-F_{y}(\tilde{X}_{i}^{m}(\frac{jr}{N}))).
  \label{eq:mac_extend_approx}
\end{split}
\end{equation}

We compute the MAC of each layer of pre-trained networks, rank them in descending order based on their MAC values, and pick top-$k\%$ layers for fine-tuning.
The networks are initialized by pre-trained weight and only top-$k\%$ layers will be fine-tuned.
More implementation details can be found in the supplementary material.

\begin{table*}[t]
  \centering
  \renewcommand\arraystretch{1.1}
  \caption{Quantitative comparison on seven challenging image restoration tasks, including dehazing, deraining, denoising, motion deblurring, low-light image enhancement (LLIE), kernel deblurring, and JPEG artifact removal.  \textbf{boldface} and \underline{underlined} indicate the best and second-best results, respectively.}
  \label{tab:allinone}
  \resizebox{\textwidth}{!}{
  \begin{tabular}{c|ccccccc|c@{}}
    \toprule
     \multirow{2}{*}{Method}
     & SOTS~\cite{sots} & Rain13k-Test\cite{degae} & BSD68~\cite{bsd68} & GoPro~\cite{gopro} & LOL~\cite{LOL} & LSDIR-Blur~\cite{li2023lsdir} & LSDIR-Jpeg~\cite{li2023lsdir} & Average \\
      & 
  PSNR$\uparrow$/SSIM$\uparrow$& PSNR$\uparrow$/SSIM$\uparrow$& PSNR$\uparrow$/SSIM$\uparrow$ & PSNR$\uparrow$/SSIM$\uparrow$ & PSNR$\uparrow$/SSIM$\uparrow$ & PSNR$\uparrow$/SSIM$\uparrow$ & PSNR$\uparrow$/SSIM$\uparrow$ & PSNR$\uparrow$/SSIM$\uparrow$  \\
    \midrule
    Restormer~\cite{Zamir2021Restormer}& 22.89/0.9172 & 27.05/0.8469 & \textbf{30.95}/\textbf{0.8657} & 27.46/0.8497 & \underline{23.65}/\underline{0.8458} & 19.60/0.3658 & \textbf{30.46}/\textbf{0.9141} & 26.01/0.8007\\
    MPRNet~\cite{mehri2021mprnet}& 25.23/0.9463 & 25.36/0.8068 & 29.83/0.8317 & 25.90/0.7949 & 22.29/0.8170 & 25.68/0.8281 & 28.96/0.8865 & 26.18/0.8445\\
     NAFNet~\cite{nafnet} & 25.74/0.9445 & 24.65/0.7877& 30.37/0.8540 & 25.53/0.7909 & 21.50/0.8104 & 29.08/0.9130 & 29.09/0.8955 & 26.57/0.8566 \\
    DL~\cite{DL} & 21.16/0.9042 & 19.56/0.6508 & 16.15/0.5861 & 17.63/0.5862 & 19.26/0.7777 & 17.98/0.6121 &19.55/0.6965 & 18.75/0.6877\\
    TAPE~\cite{liu2022tape} & 25.14/0.9319 & 23.66/0.7818 & 30.11/0.8354 & 25.97/0.7962 & 18.95/0.7632 & 24.26/0.7654 & 29.28/0.8965 & 25.34/0.8243\\
    AirNet~\cite{airnet}& 21.66/0.8366 & 20.21/0.6402 & 27.99/0.7250 & 23.36/0.7503 & 16.65/0.6708 & 23.84/0.7358 & 24.36/0.8020 & 22.58/0.7372\\
    \midrule
    SwinIR~\cite{liang2021swinir} & 27.29/0.9622 & 25.32/0.8258 & 30.65/0.8540 & 26.61/0.8125 & 18.66/0.8048 & 27.82/0.8839   & 30.13/0.9071 & 26.64/0.8643\\
    \textbf{\methodname-SwinIR} & 28.47/\underline{0.9689} & 26.31/0.8486 & 30.83/0.8611 & 26.89/0.8200 & 21.62/0.8291& 26.66/0.8514 &30.22/0.9096&27.28/0.8698\\
    \midrule
    PromptIR~\cite{promptir} & \underline{28.70}/0.9659 & \underline{27.46}/\underline{0.8585} & 30.84/\underline{0.8625} & \underline{27.71}/\underline{0.8565} & 21.19/0.8356 &  \textbf{31.01}/\textbf{0.9385} & 30.30/0.9117 & \underline{28.17}/\underline{0.8899}\\
    \textbf{\methodname-PromptIR} & \textbf{29.64}/\textbf{0.9695} & \textbf{28.47}/\textbf{0.8751}& \underline{30.86}/0.8624& \textbf{28.02}/\textbf{0.8592}& \textbf{24.46}/\textbf{0.8581} & \underline{29.57}/\underline{0.9179} & \underline{30.33}/\underline{0.9119}& \textbf{28.76}/\textbf{0.8935} \\
    \bottomrule
  \end{tabular}}
\end{table*}

\begin{figure*}[t]
  \centering
   \includegraphics[width=1\linewidth]{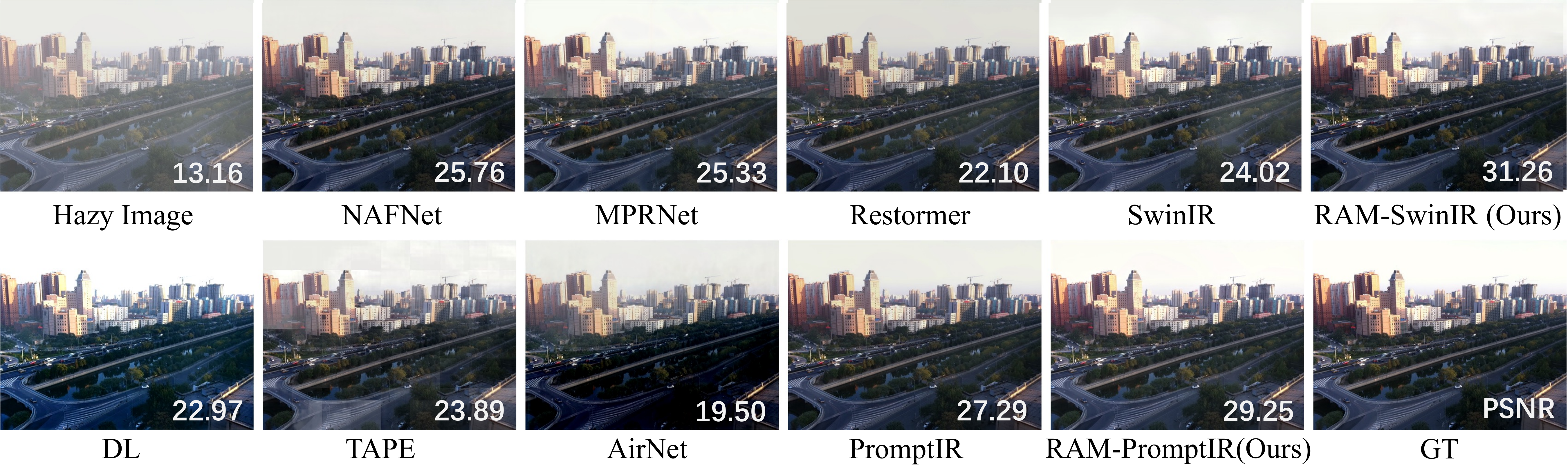}
   \caption{Dehaze visual comparison on SOTS dataset. Zoom in for details.}
   \label{fig:haze_result}
\end{figure*}

\section{Experiment}
\subsection{Experiments Settings}
\textbf{Datasets and Metrics.}  We combine datasets from various restoration tasks to form the training set, following 
 \cite{zhang2023ingredient}. For high-cost tasks that degradations are difficult to synthesize, we leverage existing paired datasets, including RESIDE\cite{sots} for dehazing, Rain13k~\cite{fu2017clearing,li2018recurrent,li2016rain,luo2015removing,yang2020single} for deraining, GoPro~\cite{gopro} for motion deblurring, and LOL-v2~\cite{lolv2} for low-light image enhancement (LLIE). For low-cost tasks that degradations are easy to synthesize (\eg noise, kernel blur, and JPEG artifact), we generate corrupted images on the LSDIR dataset~\cite{li2023lsdir} during the training process, which involves generating Gaussian noise with random variation $\sigma\in(0,50]$, creating gaussian blurred images with a blur kernel of size $k=15$ and random $\sigma\in[0.1,3.1]$, and introducing JPEG artifacts with a random quality parameter $q\in[20,90]$.

 For evaluation, we use SOTS-outdoor~\cite{sots} for dehazing, Rain13k-Test (the combination of Rain100L\cite{rain100}, Rain100H\cite{rain100}, Test100\cite{test100}, Test1200\cite{test1200} and Test2800\cite{test2800}) for deraining, GoPro for motion deblurring, LOL~\cite{LOL} for low-light enhancement, BSD68~\cite{bsd68} for denoising, LSDIR-val for kernel deblurring and jpeg artifact removal. Furthermore, We conducted evaluations including denoising tests with variances of 15, 25, and 50, deblurring tests at $k=15$ and $\sigma=2.0$, and JPEG artifact removal tests at $q=50$.

\noindent \textbf{Implementation Details.} We apply our proposed \textbf{RAM} to SwinIR~\cite{liang2021swinir} and PromptIR~\cite{promptir}. The input size for RAM-SwinIR is 64, while for RAM-PromptIR it is 128. During the pre-training phase, we use the Adam optimizer to train RAM-SwinIR and RAM-PromptIR for 300 epochs, with the learning rate decaying from 1e-4 to 6e-5 following a cosine schedule. In the fine-tuning phase, we use the Adam optimizer to fine-tune the network layers obtained from the MAC analysis of RAM-SwinIR and RAM-PromptIR for 40 epochs, with the learning rate decaying from 2e-4 to 1e-7 following a cosine schedule. The batch sizes for RAM-SwinIR and RAM-PromptIR during the pre-training and fine-tuning phases are (12,4) and (4,4), respectively.

\begin{table}[t]
\scriptsize
  \centering
  \caption{Quantitative Gaussian denoising results at different noise levels on BSD68 and Urban100 datasets in terms of PSNR.}
  \label{tab:denoise_test}
  \begin{tabular}{c|cccc|cccc@{}}
    \toprule
     \multirow{2}{*}{Method}
     & \multicolumn{4}{c}{BSD68~\cite{bsd68}}  & \multicolumn{4}{c}{Urban100~\cite{urban100}} \\
      & $\sigma=15$ & $\sigma=25$ & $\sigma=50$ & Average & $\sigma=15$ & $\sigma=25$ & $\sigma=50$& Average \\
    \midrule
    NAFNet~\cite{nafnet} & 33.22 & 30.59 & 27.30 & 30.37 & 32.67 & 30.21  & 26.97  & 29.92  \\
    MPRNet~\cite{mehri2021mprnet} & 32.73  & 30.11  & 26.65  & 29.83  & 32.06  & 29.46  & 25.77  & 29.10  \\    
    Restormer~\cite{Zamir2021Restormer} & 33.79  & 31.17  & 27.90  & 30.95  & 33.83 & 31.40  & 27.99  & 31.07  \\
    DL~\cite{DL} & 16.04  & 16.20  & 16.19  & 16.15  & 19.17  & 19.11  & 18.47  & 18.92 \\
    TAPE~\cite{liu2022tape} & 33.10  & 30.37  & 26.86  & 30.11  & 32.59  & 29.93  & 26.19  & 29.57 \\
    AirNet~\cite{airnet} & 31.63  & 28.83  & 23.52  & 27.99  & 29.79  &26.90  & 21.35  & 26.01 \\
    \midrule
    SwinIR~\cite{liang2021swinir} & 33.53  & 30.89  & 27.54  & 30.65  & 33.50  & 30.99  & 27.37  & 30.62 \\
    {\methodname-SwinIR} & {33.65} & {31.06} & {27.77}  & {30.82}  & {33.82}  & {31.43} & {27.94}  & {31.07}  \\
     \rowcolor{mygray} {performance gains} & \bonus{($\uparrow$0.12)}  & \bonus{($\uparrow$0.17)}  & \bonus{($\uparrow$0.23)}  & \bonus{($\uparrow$0.17)}  & \bonus{($\uparrow$0.32)}  & \bonus{($\uparrow$0.44)}  & \bonus{($\uparrow$0.57)}  & \bonus{($\uparrow$0.45)} \\
    \midrule
    PromptIR~\cite{promptir} & 33.67  & 31.06  & 27.80  & 30.84  & 33.56  & 31.08  & 27.64  & 30.76  \\
    {\methodname-PromptIR} & {33.70}  & {31.08}  & {27.79}  & {30.86}  & {33.70}  & {31.30}  & {27.92}  & {30.97} \\
    \rowcolor{mygray} {performance gains}& \bonus{($\uparrow$0.03)}  & \bonus{($\uparrow$0.02)}  & \decline{($\downarrow$0.01)}  & \bonus{($\uparrow$0.02)}  & \bonus{($\uparrow$0.14)}  & \bonus{($\uparrow$0.22)}  & \bonus{($\uparrow$0.28)}  & \bonus{($\uparrow$0.21)} \\
    \bottomrule
  \end{tabular}
\end{table}

\subsection{Comparisons}
To validate the gain capability and effectiveness of our \methodname, we apply the proposed RAM to SwinIR (a general image restoration method) and PromptIR (an all-in-one image restoration method). Four general architecture-based image restoration methods \cite{nafnet,mehri2021mprnet,Zamir2021Restormer,liang2021swinir} and four all-in-one methods \cite{DL,liu2022tape,airnet,promptir} are considered for comparison. We ensure that the number of supervised pixels employed by all other methods equals that used during the pre-training stage.

As illustrated in ~\cref{tab:allinone}, our approach achieves the best or comparable performance on each task.
On the average score across seven different tasks, our method with PromptIR~\cite{promptir} achieves $0.59$dB performance gains compared to the second-best algorithm.
Besides, the SwinIR equipped with \methodname{} also yields $2.40\%$ improvement on PSNR.
Specifically, our \methodname{} has significant benefits for dehazing and low-light enhancement. \cref{tab:denoise_test} shows the Quantitative denoising result at different noise levels. Both \methodname-SwinIR and \methodname-PromptIR get higher performance than the origin versions.

\begin{figure*}[t]
  \centering
   \includegraphics[width=1\linewidth]{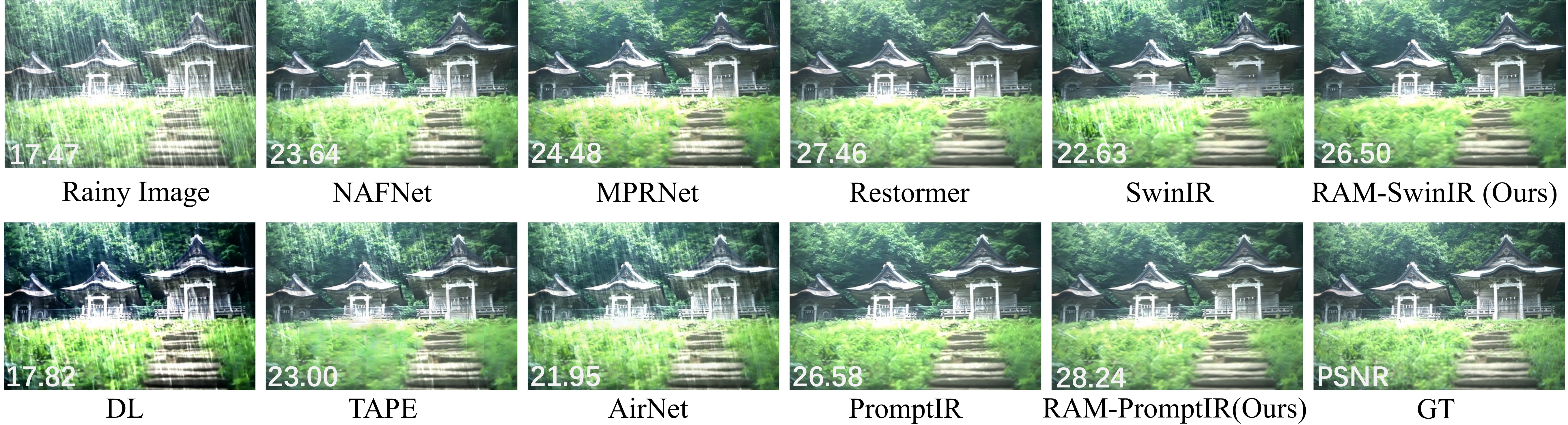}
   \caption{Derain visual comparsion on Rain13k-Test dataset. Zoom in for details.}
   \label{fig:rain_result}
\end{figure*}

\begin{figure*}[t]
  \centering
   \includegraphics[width=1\linewidth]{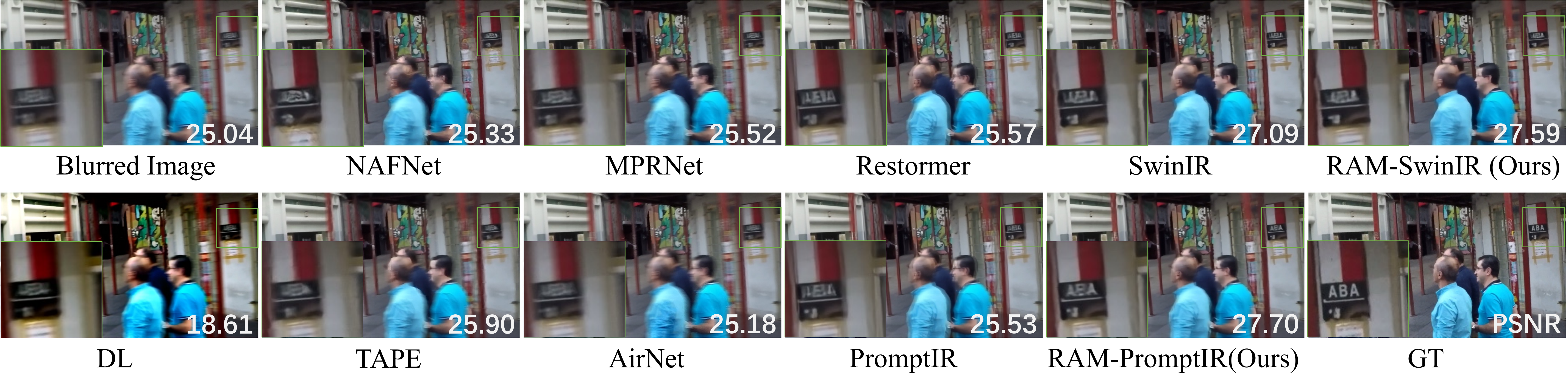}
      \setlength{\abovecaptionskip}{2pt}
   \caption{Motion deblur visual comparison on GoPro dataset. Zoom in for details.}
   \label{fig:blur_result}
\end{figure*}

\begin{figure*}[t]
  \centering
   \includegraphics[width=1\linewidth]{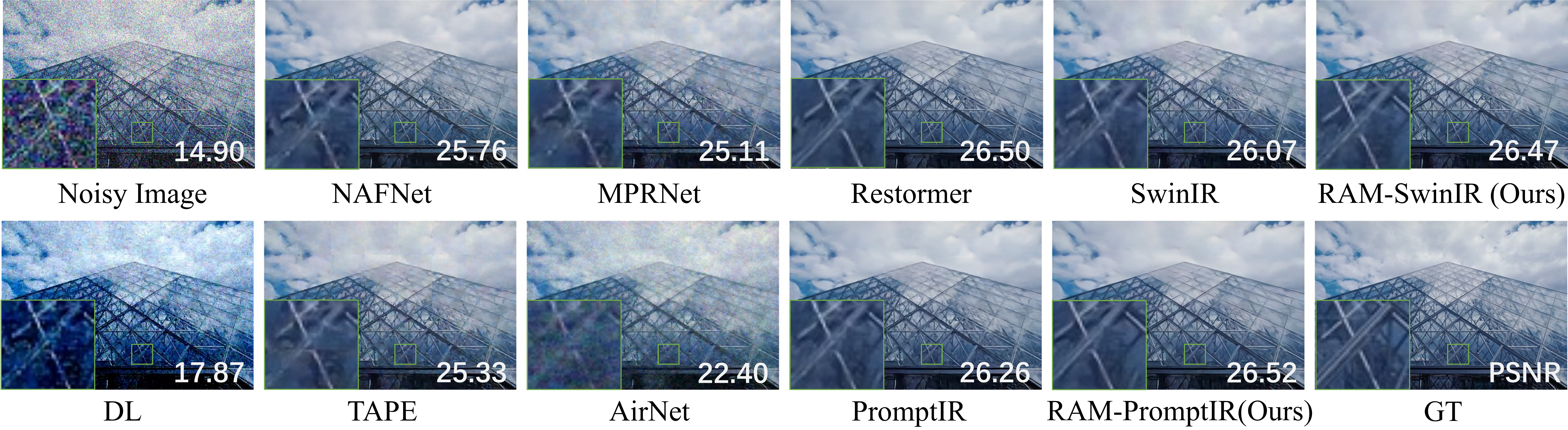}
      \setlength{\abovecaptionskip}{2pt}
   \caption{Denoising visual comparison on CBSD68 dataset. Zoom in for details.}
   \label{fig:noise_result}
\end{figure*}

\begin{figure*}[t]
  \centering
   \includegraphics[width=1\linewidth]{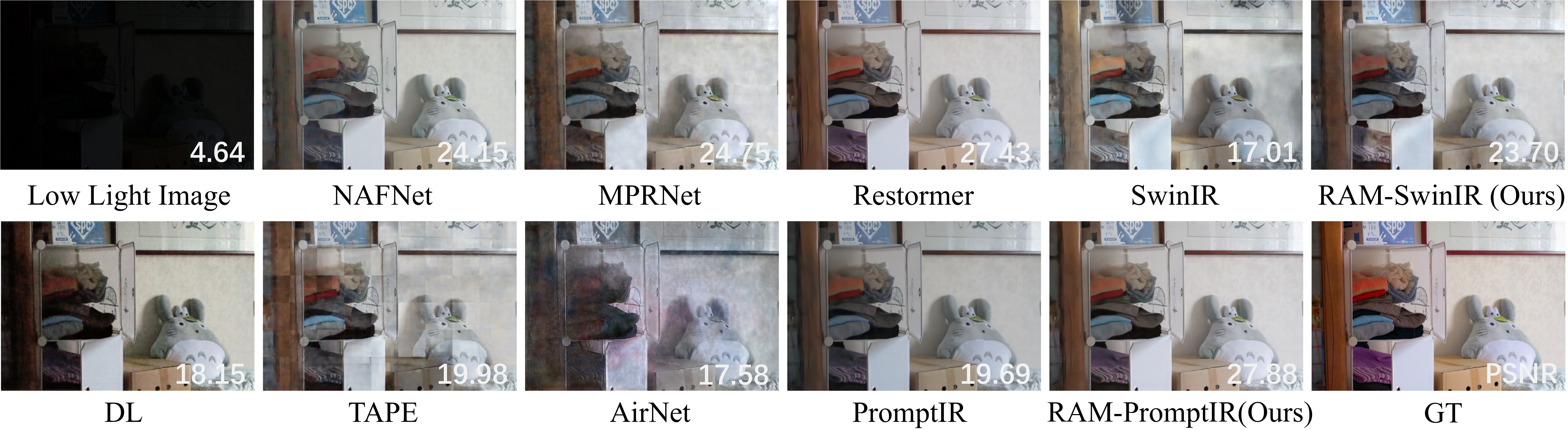}
   \caption{LLIE visual comparison on LOL dataset. Zoom in for details.}
   \label{fig:lowlight_result}
\end{figure*}

~\cref{fig:haze_result}-\cref{fig:lowlight_result} show the qualitative results of various methods on different datasets. In \cref{fig:haze_result}, our method achieves better dehazing effects (right region) and exposure correction (sky). In the deraining task (\cref{fig:rain_result}, our method better removes rain streaks and restores textures in the occluded regions. In terms of denoising (\cref{fig:noise_result}) and deblurring (\cref{fig:blur_result}), we achieve clearer results with fewer artifacts. We also demonstrate better color correction (the purple blanket on the left) and exposure correction in low-light image enhancement tasks (\cref{fig:lowlight_result}). For simplicity, the qualitative effects of kernel deblurring and JPEG artifact removal will be presented in the supplementary material.

\subsection{Ablation Study}
\label{sec:ablation}
In this section, we conduct an ablative study on the masking ratio, mask patch size, pre-training strategy, fine-tuning strategy, and fine-tuning ratio to demonstrate the effectiveness of our MIM pre-training and fine-tuning strategy.
\begin{table}[H]
    \centering
    \setlength{\tabcolsep}{10pt}
    \small
    \caption{\textbf{Ablative results on masking ratios.}}
    \resizebox{0.7\linewidth}{!}{
    \begin{tabular}{@{}c|ccccc@{}}
    \toprule
    Masking ratio  & $20\%$ & $40\%$ & $50\%$ & $60\%$ & $80\%$ \\ \midrule
    PSNR$\uparrow$ & 27.28  & 27.21  & \textbf{27.28}  & 27.26  & 27.08  \\
    SSIM$\uparrow$ & 0.8663 & 0.8683 & \textbf{0.8698} & 0.8694 & 0.8642 \\ \bottomrule
    \end{tabular}}
    \label{tab:mask_ratio_ablation}
\end{table}

\begin{table}[t]
    \begin{minipage}[c]{0.48\textwidth}
        \scriptsize
        \setlength{\tabcolsep}{3.8pt}
        \caption{
            \textbf{Ablative results of different pre-training strategies.}
        }
        \renewcommand{\arraystretch}{1.4}
        \begin{tabular}{l|ccccc}
        \toprule
        RAM-SwinIR & PSNR$\uparrow$ & SSIM$\uparrow$\\ \midrule
        pre-trained w/ gt & 26.62 & 0.8580  \\
        pre-trained w/ paired data & 27.28 & 0.8698 \\
 \bottomrule
        \end{tabular}
        \label{tab:pretrain_strategy_ablation}
    \end{minipage}
    \hfill
    \begin{minipage}[c]{0.48\textwidth}
        \scriptsize
        \setlength{\tabcolsep}{14pt}
        \caption{
            \textbf{Ablative results of different fine-tuning strategies.}
        }
        \renewcommand{\arraystretch}{1.4}
      \begin{tabular}{@{}c|cc@{}}
        \toprule
        RAM-SwinIR & PSNR$\uparrow$ & SSIM$\uparrow$\\ \midrule
        random & 26.86 & 0.8535  \\
        IG\cite{ig} & 26.92 & 0.8554  \\
        MAC (Ours) & \textbf{27.28} & \textbf{0.8698} \\ \bottomrule
        \end{tabular}
        \label{tab:finetune_strategy_ablation}
    \end{minipage}
\end{table}
\noindent\textbf{Patch size \& masking ratio}
are two essential hyper-parameters that determine the continuity and area of the masking of an image.
In high-level tasks, MAE~\cite{mae} masks $75\%$ of an image with $16 \times 16$ patch size.
However,  it can corrupt the local details of images, which is not suitable for image restoration.

\begin{wraptable}{l}{7.2cm}
  \caption{\textbf{Ablative results in terms of the PSNR on fine-tuning ratios.} We compared the performance in restoring images with unseen noises (Out-of-Distribution Denoising) and known degraded images (In-Distribution). In this case, the settings of In-Distribution are the same as \cref{tab:allinone}.}
  \renewcommand\arraystretch{1.25}
  \resizebox{\linewidth}{!}{
  \begin{tabular}{@{}c|cccc|c@{}}
    \toprule
    \small
    & \multicolumn{4}{c}{Out-of-Distribution Denoising} & In-Distribution\\
    Method & Possion & Pepper & Speckle & Average & Average\\
    \midrule
    SwinIR~\cite{liang2021swinir} & 12.83 & 10.00 &20.86 &14.56&26.64 \\
    RAM-SwinIR$_{10\%}$ & \textbf{13.67}  & \textbf{19.23} & \textbf{21.07} & \textbf{17.99} & 27.28\\
    RAM-SwinIR$_{20\%}$ & 13.27  & 19.09 & 20.68 & 17.68 & 27.35\\
    RAM-SwinIR$_{50\%}$ & 12.75  &  16.51 & 20.36 & 16.54& 27.38\\
    RAM-SwinIR$_{100\%}$ & 12.47  &  15.31 & 20.01 & 15.93 & \textbf{27.54}\\
    \bottomrule
  \end{tabular}}
  \label{tab:finetune_ratio_ablation}
\end{wraptable}

We first find the best choice of patch size by pre-training SwinIR~\cite{liang2021swinir} on $1\times 1$, $4\times4$, and $8\times8$, as shown in ~\cref{fig:patch-size}.
Since the attention layers of SwinIR treat an $8\times8$ patch as a token, the $4\times4$ pre-training produces heavy artifacts.
Besides, the results generated by $8 \times 8$ pre-training are highly missing details, \eg the texture of the polar bear's paws.
In contrast, the model pre-trained with $1\times1$ patch size, which is also our final choice, achieves a satisfactory reconstruction and removes most of the rain streaks.

Then, we adjust the masking ratio from $20\%$ to $80\%$. As we can see in ~\cref{tab:mask_ratio_ablation}, the model pre-trained with $50\%$ achieves the highest performance.
Moreover, the performance is significantly dropped from $27.28$dB to $27.08$dB in terms of PSNR when we continue to increase the masking ratio, which also demonstrates our opinion that a high masking ratio is harmful to image restoration.

\noindent\textbf{Pre-trained with paired data.} 
\cref{tab:pretrain_strategy_ablation} compares the results of using paired data for mask image pretraining (our pretraining strategy) with those using only ground truth for mask image pretraining. It shows that pre-trained with paired data is necessary for our RAM.
Pretraining the model on high-quality images does not effectively enable learning for image restoration tasks. It still requires paired data to guide the model in the learning process.

\noindent\textbf{Fine-tuning strategy.}
To verify the effectiveness of our fine-tuning strategy, we fine-tune $10\%$ of the network layers selected through MAC analysis, IG\cite{ig}, and uniform sampling, respectively, and the results are shown in \cref{tab:finetune_strategy_ablation}. Compared to IG, we have improved by 0.36 dB in PSNR and 1.6\% in SSIM, which indicates that our selection strategy is superior to IG.

\noindent\textbf{Fine-tune ratio.} 
We conduct the ablation experiment to compare the network's performances with different fine-tune ratios in \cref{tab:finetune_ratio_ablation}. We found that using our finetune strategy, a pre-trained network could achieve comparable performance by fine-tuning only a few layers (\eg $10\%$). At the same time, we need to fine-tune almost all network parameters to get the best performance on given tasks.

\noindent\textbf{Performance \textit{vs} Generalization capability.} 
We found a trade-off between in-distribution performance and out-of-distribution generalization in~\cref{tab:finetune_ratio_ablation}. We found that the more layers fine-tuned, the less generalization capability to tackle the out-of-distribution tasks. With our fine-tuning method, the model can have stronger generalization while maintaining comparable performance.

\section{Conclusion}
This paper presents \methodname{}, a pipeline for extracting intrinsic image information from corrupted images using Mask Image Modeling (MIM) pre-training. 
We design a MIM pre-training strategy tailored for image restoration and a fine-tuning algorithm to handle the transition from masked to complete images.
By analyzing layer importance with MAC, we achieve high performance with minimal parameter tuning.
Extensive experiments demonstrate that our \methodname{} can bring boosts to various architectures and achieve state-of-the-art performance, moving towards a unified solution for all-in-one image restoration.
\bibliographystyle{splncs04}
\bibliography{main}

%
%
\clearpage
\appendix
\title{Supplementary Material of Restore Anything with Masks:  Leveraging Mask Image Modeling for Blind All-in-One Image Restoration} 

\author{
Chu-Jie Qin\inst{1,2}\thanks{A part of this work is done during Chu-Jie Qin's internship at Samsung.} \and
Rui-Qi Wu\inst{1,2} \and
Zikun Liu\inst{3} \and 
Xin Lin\inst{5} \and
Chun-Le Guo\inst{1,2} \and
Hyun Hee Park\inst{4} \and
Chongyi Li\inst{1,2}\thanks{Chongyi Li is the corresponding author.}
}

\titlerunning{Restore Anything with Masks}

\institute{VCIP, CS, Nankai University  \and
NKIARI, Shenzhen Futian \\
\email{\{chujie.qin,wuruiqi\}@mail.nankai.edu.cn}\\
\email{\{guochunle, lichongyi\}@nankai.edu.cn}
\and
Samsung Research, China, Beijing (SRC-B)\and
The Department of Camera Innovation Group, Samsung Electronics \\
\email{\{zikun.liu,inextg.park\}@samsung.com} \and
Sichuan University  \\
\email{linxin@stu.scu.edu.cn}\\
}
\maketitle
\begin{table}[t]
  \centering
  \renewcommand\arraystretch{1.1}
  \caption{Quantitative comparison on Rain13k-Test, which consists of Rain100L~\cite{rain100}, Rain100H~\cite{rain100}, Test100~\cite{test100}, Test1200~\cite{test1200}, and Test2800~\cite{test2800}.  \textbf{Boldface} and \underline{underlined} indicate the best and second-best results, respectively.}
  \label{tab:rain_test}
  \resizebox{\textwidth}{!}{
  \begin{tabular}{c|ccccc|c@{}}
    \toprule
     \multirow{2}{*}{Method}
     & Rain100L~\cite{rain100} & Rain100H~\cite{rain100} & Test100~\cite{test100} & Test1200~\cite{test1200} & Test2800~\cite{test2800} & Average \\
    \vspace{0.5pt}
      & 
  PSNR$\uparrow$/SSIM$\uparrow$& PSNR$\uparrow$/SSIM$\uparrow$& PSNR$\uparrow$/SSIM$\uparrow$ & PSNR$\uparrow$/SSIM$\uparrow$ & PSNR$\uparrow$/SSIM$\uparrow$ & PSNR$\uparrow$/SSIM$\uparrow$  \\
    \midrule
    NAFNet~\cite{nafnet}  &  25.83/0.8509 & 19.02/0.6141 & 22.46/0.7729& 27.90/0.8383& 28.02/0.8624 & 24.65/0.7877 \\
    MPRNet~\cite{mehri2021mprnet} & 25.34/0.8381 & 21.92/0.7062 & 22.51/0.7742 & 28.24/0.8383 & 28.78/0.8770 & 25.36/0.8068 \\
    Restormer~\cite{Zamir2021Restormer} & \textbf{28.49}/\textbf{0.8887} & \underline{25.98}/\textbf{0.8326} & 22.58/0.7598 & 28.13/0.8464 & 30.06/0.9069 & 27.05/0.8469\\
    DL~\cite{DL}  & 20.59/0.7453 & 13.62/0.3649 & 18.79/0.6251 & 21.84/0.7474 & 22.93/0.7712 & 19.56/0.6508 \\
    TAPE~\cite{liu2022tape}  & 23.67/0.8135 & 17.22/0.6135 & 22.22/0.7750 & 27.56/0.8413 & 27.61/0.8665 & 23.66/0.7818 \\
    AirNet~\cite{airnet} & 19.66/0.6697 & 14.32/0.3957 & 20.70/0.6595 & 23.37/0.7400 & 23.02/0.7359 & 20.21/0.6402 \\
    \midrule
    SwinIR~\cite{liang2021swinir} & 23.85/0.8216 & 21.88/0.7579 & 23.19/0.7903 & 28.28/0.8608 & 29.43/0.8983 & 25.32/0.8258 \\
    \textbf{\methodname-SwinIR} & 26.08/0.8718 & 23.61/0.7931& 23.47/0.8085& 28.56/0.8663& 29.84/0.9033&26.31/0.8486\\
    \midrule
    PromptIR~\cite{promptir} & 26.83/0.8609 & 25.29/0.8181 & \underline{24.70}/\underline{0.8254} & \underline{29.59}/\underline{0.8746} & \underline{30.89}/\underline{0.9135} & \underline{27.46}/\underline{0.8585} \\
    \textbf{\methodname-PromptIR} & \underline{28.12}/\underline{0.8855}& \textbf{26.36}/\underline{0.8314}& \textbf{25.61}/\textbf{0.8510}& \textbf{30.99}/\textbf{0.8915}& \textbf{31.29}/\textbf{0.9163}& \textbf{28.47}/\textbf{0.8751} \\
    \bottomrule
  \end{tabular}}
\end{table}
\begin{figure}[t]
  \centering
   \includegraphics[width=1\linewidth]{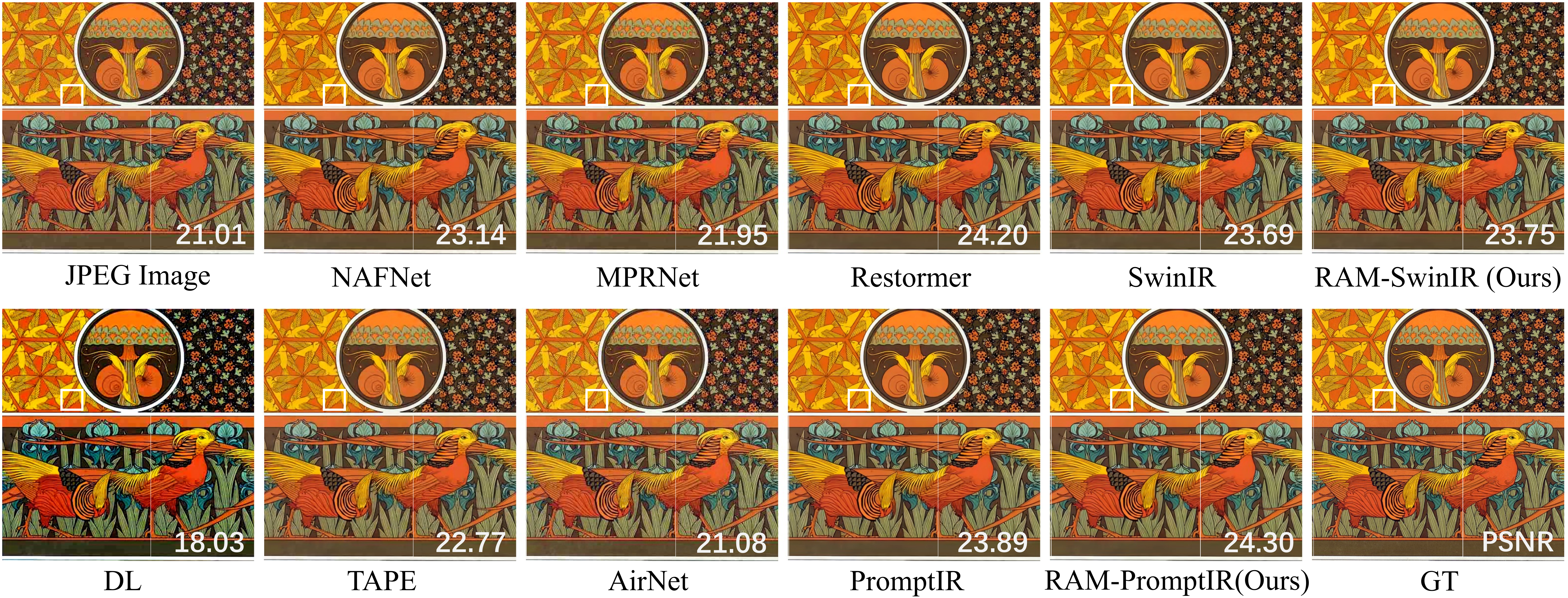}
      \setlength{\abovecaptionskip}{2pt}
   \vspace{-0.4cm}
   \caption{JPEG artifact removal comparison on LSDIR\cite{li2023lsdir} dataset. Zoom in for details.}
   \label{fig:jpeg1}
\end{figure}

\begin{figure}[t]
  \centering
   \includegraphics[width=1\linewidth]{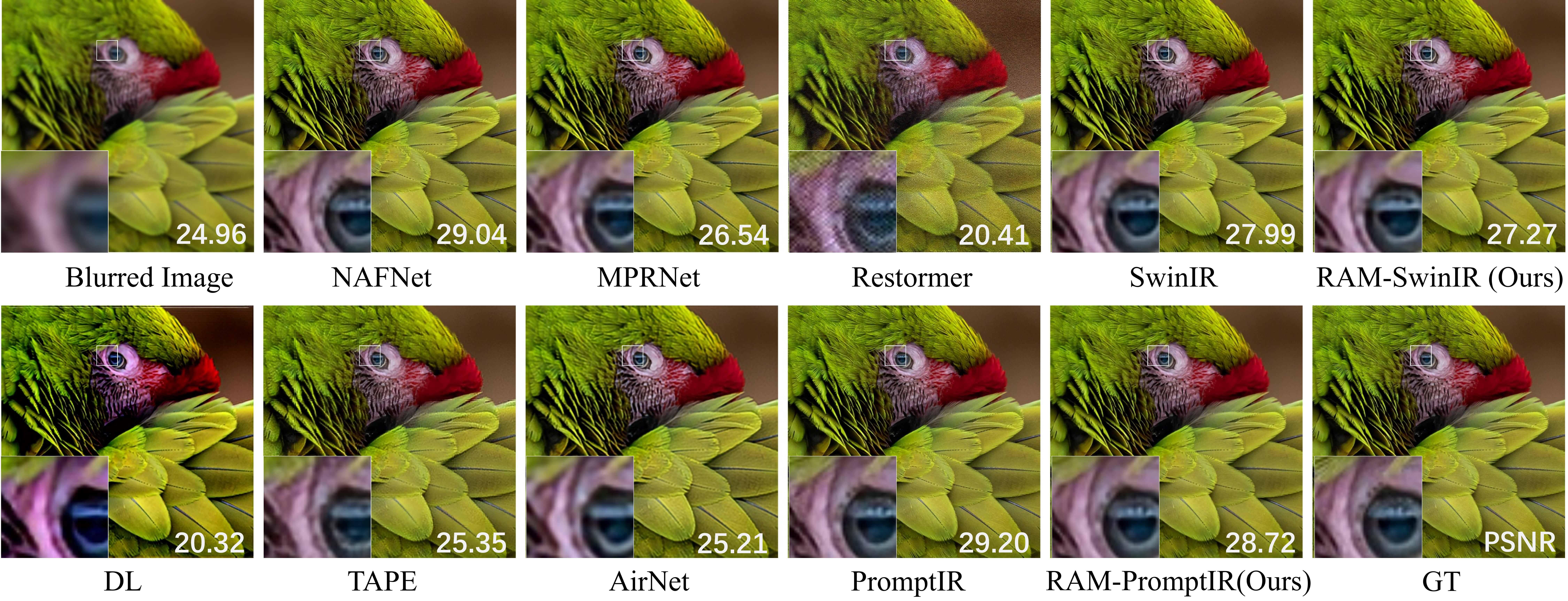}
      \setlength{\abovecaptionskip}{2pt}
   \vspace{-0.4cm}
   \caption{Kernel deblur comparison on LSDIR\cite{li2023lsdir} dataset. Zoom in for details.}
   \label{fig:kernelblur1}
   \vspace{-0.4cm}
\end{figure}

\begin{figure}[t]
  \centering
   \includegraphics[width=1\linewidth]{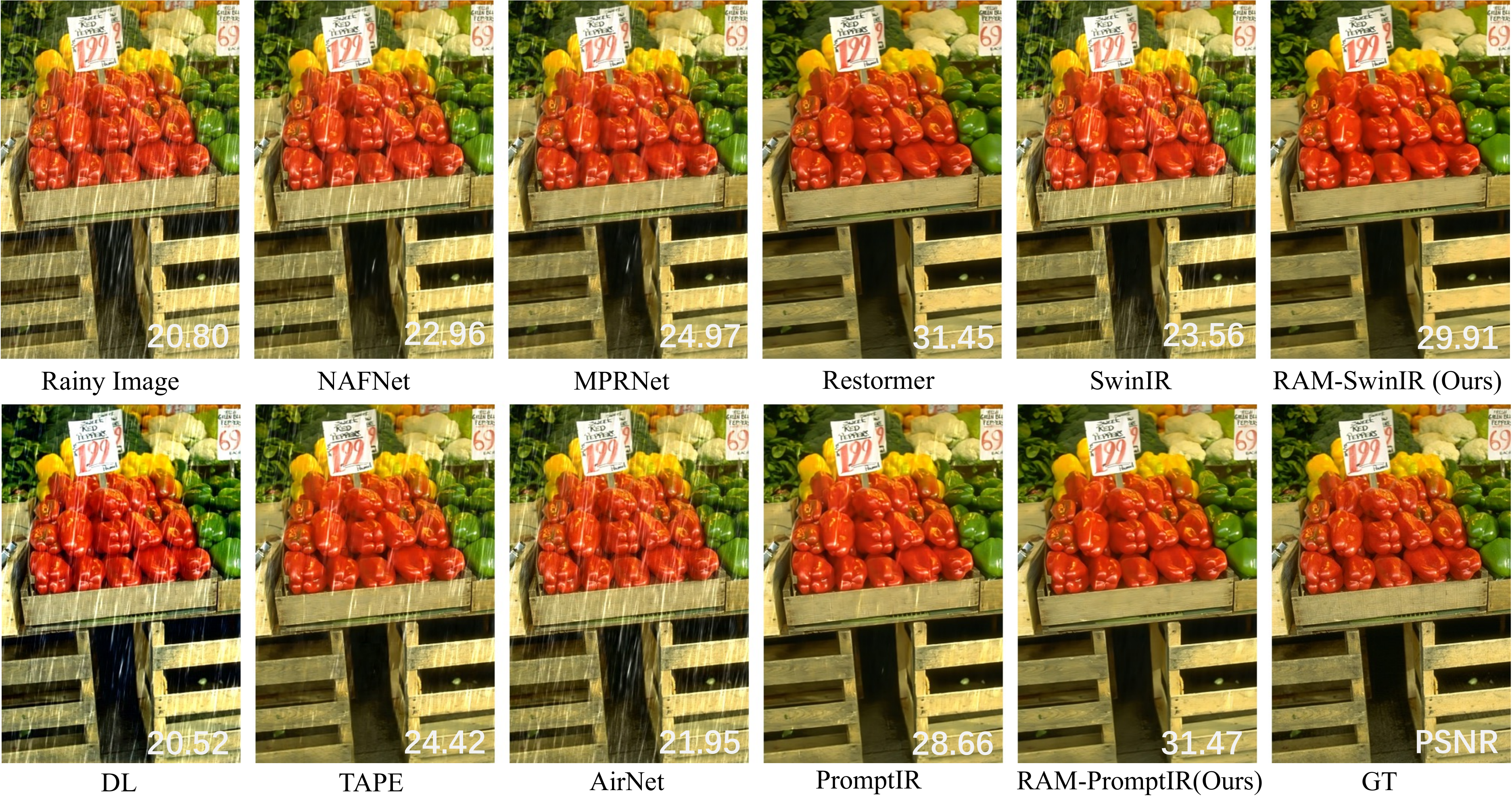}
      \setlength{\abovecaptionskip}{2pt}
   \vspace{-0.15cm}
   \caption{Derain comparison on Rain13k-Test\cite{rain100,test100,test1200,test2800} dataset. Zoom in for details.}
   \label{fig:rain2}
   \vspace{-0.25cm}
\end{figure}

\section{Addtional Details}
This section primarily provides additional implementation details not covered in the main text, including low-cost degradation synthesis (in \cref{sec:synthesis}) and details of mask attribute conductance (MAC) analysis (in \cref{sec:map}).
\subsection{Low-Cost degradation synthesis}
\label{sec:synthesis}
Low-cost degradation refers to degradations that can be easily synthesized. In our experimental setup, three types of low-cost degradations were involved: noise, kernel blur, and JPEG artifact. We obtained paired data for these three degradations through online synthesis during the training time. Here, we provide additional details on the specific synthesis process for each type of degradation.

\noindent{}\textbf{Gaussian Noise.} We randomly sample gaussian noise $N$ from the gaussian distribution $\mathcal{N}(0,\sigma^2)$. Subsequently, we add this Gaussian noise $N$ to the original image $I$ to obtain a noisy image $I_N$. To ensure data correctness, we truncate values that fall outside the data range:
\begin{equation}
    I_N = \mathrm{Clip}(I+N)
  \label{eq:noise_syn}
\end{equation}
Here $\mathrm{Clip}(\cdot)$ involves truncating data to the minimum or maximum value when it falls below the minimum or exceeds the maximum.

\noindent{}\textbf{Kernel Blur.} We employ a gaussian blur approach~\cite{gaussianblur} to synthesize kernel-blur degradation. By specifying the kernel size $k$ and standard deviation $\sigma$, we obtain a Gaussian blur filter $G$. Subsequently, convolving this filter with the original image generates the kernel-blurred image $I_B$:
\begin{equation}
    I_B = I\ast G
  \label{eq:blur_syn}
\end{equation}
Here, $k$ is set to 15 and $\sigma$ is randomly sampled in $[0.1,3.1]$ for training.

\noindent{}\textbf{JPEG Artifact.} 
JPEG artifacts, also known as JPEG compression artifacts, are blocky distortions that occur when an image is compressed using the lossy JPEG format. The severity of JPEG artifacts varies based on the quality $q$ of the JPEG compression applied. Therefore, we randomly applied JPEG compression to the images at different qualities (sampled in $[20,90]$), resulting in corrupted images $I_J$with varying degrees of compression artifacts:
\begin{equation}
    I_J = \mathbf{JPEG}(I;q)
  \label{eq:jpeg_syn}
\end{equation}

\subsection{Detials of MAC Analysis}
\label{sec:map}
MAC analysis is a \textbf{gradient-based} attribution method used to measure the sensitivity of various network layers to the change from masked input to whole input. We believe that network layers more sensitive to this change should undergo fine-tuning. In the main text of our paper, we refer to this sensitivity as layer importance. Considering that gradient-based attribution methods are seldom applied in the low-level domain, we provide additional explanations and specific implementation details.

\noindent{}\textbf{Mask Attribute Path.} Methods\cite{ig,approx_ig,layerconductance,intinf} based on integrated gradient attribution often require specifying an integration path, which describes the process of input changes. For example, Integrated Gradients (IG)\cite{ig} aims to attribute the impact of the original input, defining a linear path (which we refer to $\gamma(\alpha)$ in our paper) from an all-black image to the input image. In this paper, we aim to attribute the impact of changes from masked input to whole input. Therefore, we define a differentiable path that gradually reduces the masking rate along the path, \ie mask attribute path.

\noindent{}\textbf{Details of MAC.} For ease of explanation, we will copy Eq.(8) from the main text in our paper as follows:
\begin{equation}
  \mathrm{MAC}_{r}^{y}(x)\approx  \sum_{i}\int_{1-r}^{1} \frac{\partial F( \tilde{X}_{i}^{m}(\alpha;\alpha_i))}{\partial y}\cdot \frac{\partial y}{\partial \alpha}\, d\alpha.
  \label{eq:mac_extend}
\end{equation}
We adopt the expression from previous gradient-based attribution methods\cite{layerconductance,shrikumar2018computationally}, describing y as a hidden neuron.
This expression might not be sufficiently clear. Certainly, $y$ can be understood as the intermediate output obtained through hidden neurons, \ie $F_y(x)$. Therefore, it can be understood as the gradient of the network through a specific unit, which can be a neuron, a layer, or even an activation function.

\noindent{}\textbf{Hyperparameters.} From Eq.(6) and Eq.(9) in the paper, the hyperparameters to be determined include $\{\alpha_i\}$, $\delta$, $r$, and $N$. In practice, $\{\alpha_i\}$ represents a shuffled arrangement of $H\times W$  equidistant points from $[0,1]$, where $\delta$, $r$, and $N$ are set to be $10000$, $0.5$, and $200$ respectively.

\noindent{}\textbf{Sampling.} 
To more accurately evaluate the MAC of each network layer, we uniformly sampled several images containing different types of degradation for attribution analysis. More precisely, for each degradation in our settings, we randomly sampled $10$ images and computed the mean across all samples.

\begin{figure}[t]
  \centering
   \includegraphics[width=1\linewidth]{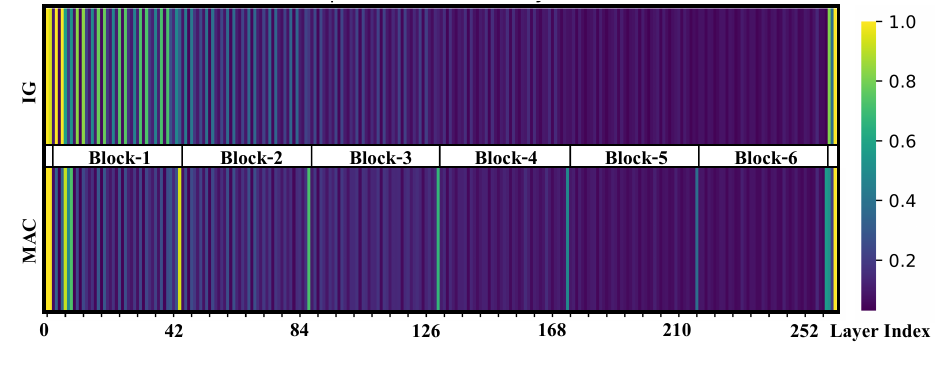}
      \setlength{\abovecaptionskip}{2pt}
   \vspace{-0.6cm}
   \caption{Importance of each layer in RAM-SwinIR.}
   \label{fig:mac}
   \vspace{-0.5cm}
\end{figure}

\noindent{}\textbf{Statistical result for MAC.} 
\cref{fig:mac} displays the heat maps of the importance of each layer in RAM-SwinIR using IG and MAC methods. Generally, the importance of the layer decreases with depth. In detail, MAC highlights the last conv layer of each transformer block, while IG doesn't. It makes sense to adjust features at the end of each block to solve the distribution shift caused by input integrity.

\begin{figure}[t]
  \centering
   \includegraphics[width=1\linewidth]{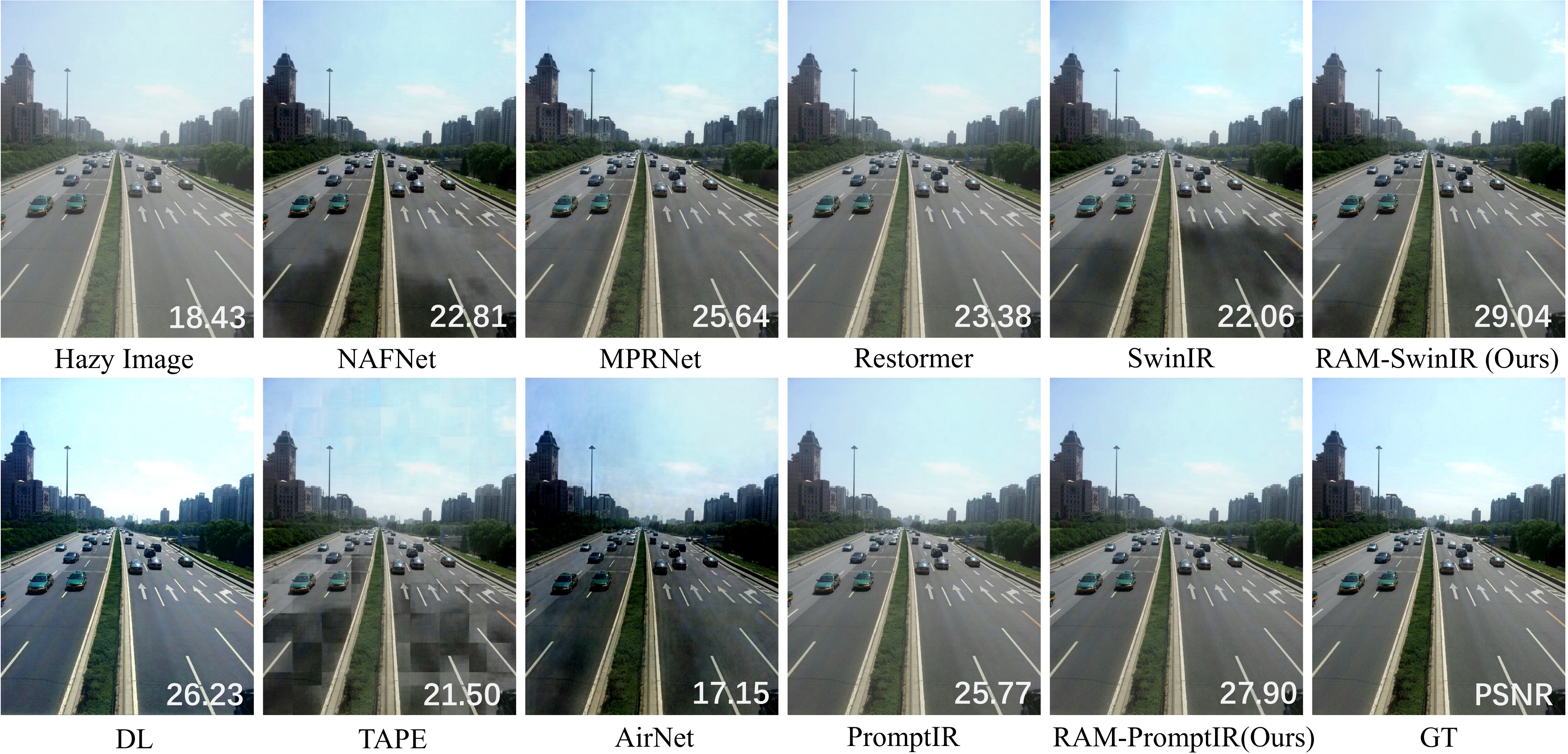}
      \setlength{\abovecaptionskip}{2pt}
   \vspace{-0.4cm}
   \caption{Dehaze visual comparison on SOTS\cite{sots} dataset. Zoom in for details.}
   \label{fig:dehaze2}
   \vspace{-0.1cm}
\end{figure}

\begin{figure}[t]
  \centering
   \includegraphics[width=1\linewidth]{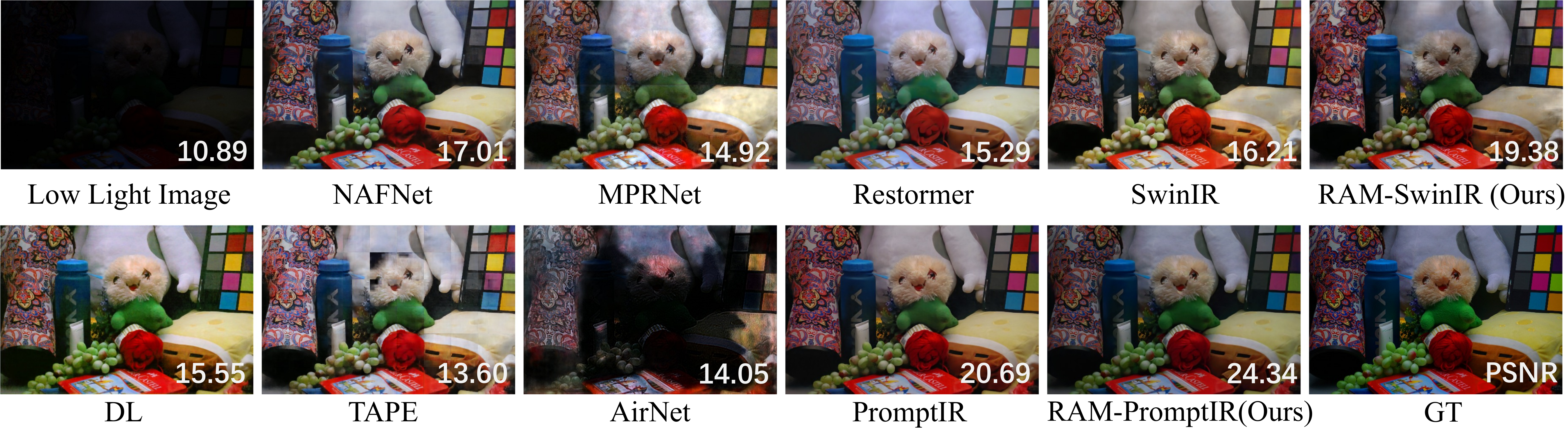}
      \setlength{\abovecaptionskip}{2pt}
   \vspace{-0.15cm}
   \caption{Low light enhancement visual comparison on LOL\cite{LOL} dataset. Zoom in for details.}
   \label{fig:lowlight2}
   \vspace{-0.25cm}
\end{figure}

\section{Additional Results}
\subsection{More quantitative results}
\noindent{}\textbf{Deraining.} \cref{tab:rain_test} shows the detailed results on Rain13-Test datasets, which is collected from Rain100L \cite{rain100}, Rain100H \cite{rain100}, Test100 \cite{test100}, Test1200 \cite{test1200}, and Test2800 \cite{test2800}. In Tab.1 of the main text of our paper, the results for the \textit{Rain13k-Test} column are also the average values obtained from the results of these five sub-test sets. It can be observed that SwinIR and PromptIR equipped with our RAM show performance improvements on almost all the deraining test sets. Furthermore, they achieved average improvements of 0.99dB and 1.01dB, respectively on these five datasets.

\vspace{-0.5cm}
\subsection{More qualitative results}
\noindent{}\textbf{Kernel Deblurring $\&$ JPEG Artifact Removal.} 
\cref{fig:jpeg1} and \cref{fig:kernelblur1} are the results for JPEG artifact removal and kernel deblurring, respectively, as mentioned in the main text. While achieving state-of-the-art performance on all other degradations, our performance on kernel-blurred images can also reach comparable performance.

\begin{figure}[t]
  \centering
   \includegraphics[width=1\linewidth]{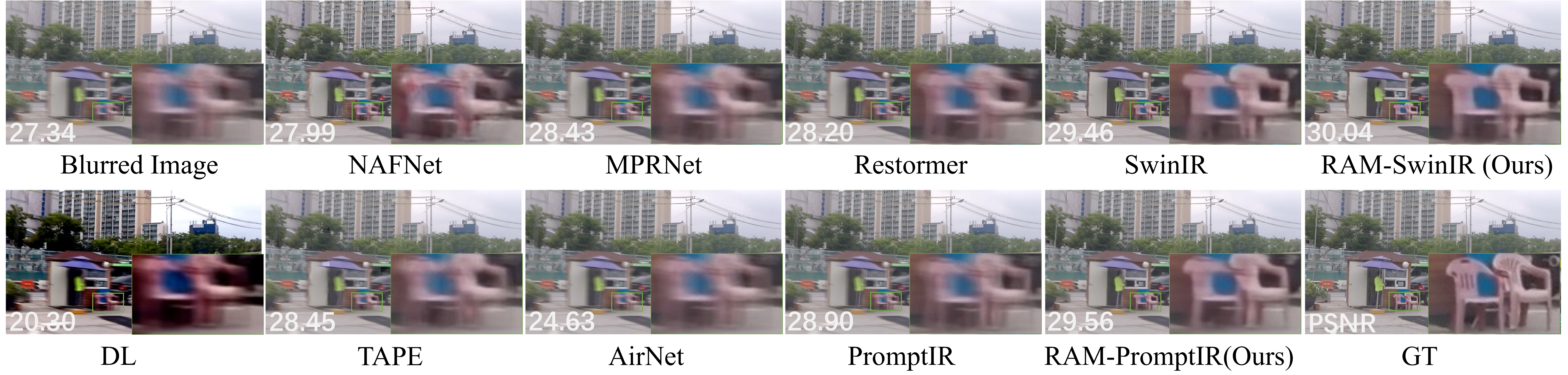}
      \setlength{\abovecaptionskip}{2pt}
   \caption{Motion Deblur visual comparison on GoPro\cite{gopro} dataset. Zoom in for details.}
   \label{fig:motiondeblur2}
\end{figure}

\begin{figure}[t]
  \centering
   \includegraphics[width=1\linewidth]{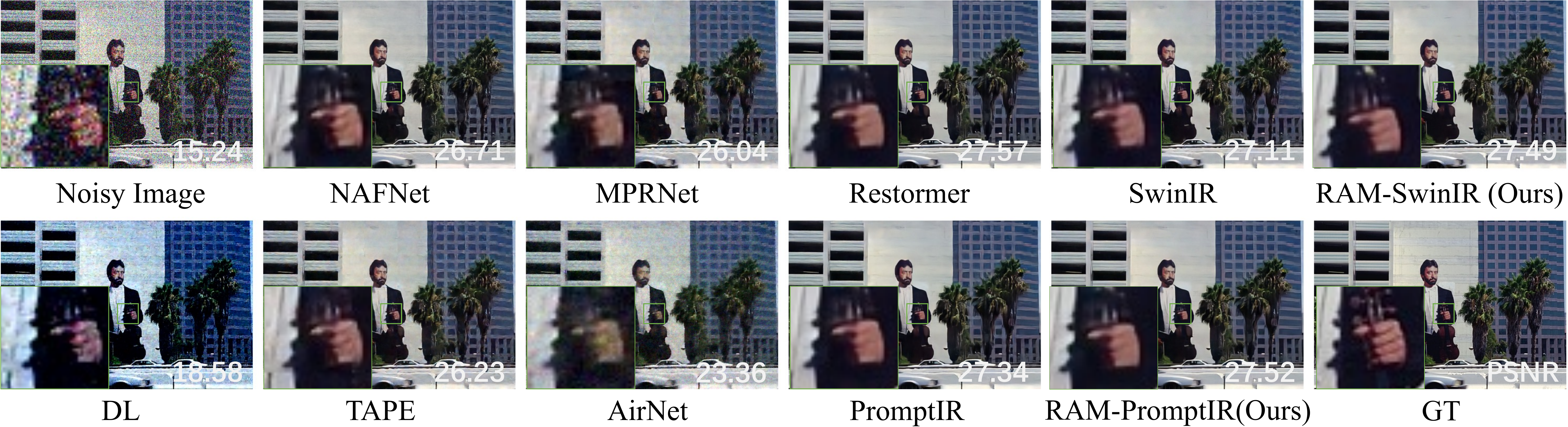}
      \setlength{\abovecaptionskip}{2pt}
   \caption{Denoising visual comparison on CBSD68\cite{bsd68} dataset. Zoom in for details.}
   \label{fig:denoise2}
\end{figure}

\noindent{}\textbf{More Results.} We provide more visual comparison on several restoration tasks in \cref{fig:rain2} to \cref{fig:denoise2}. The visual results indicate that our model significantly outperforms other methods in color correction and texture restoration.

\end{document}